\pdfoutput=1

\documentclass[11pt]{article}

\usepackage[preprint]{acl}

\usepackage{times}
\usepackage{latexsym}
\usepackage{longtable}
\usepackage{xcolor}
\usepackage{array}
\usepackage[T1]{fontenc}

\usepackage[utf8]{inputenc}

\usepackage{microtype}

\usepackage{inconsolata}

\usepackage{graphicx}

\title{\textsc{WhoDunIt}: Evaluation benchmark for culprit detection in mystery stories}

\author{Kshitij Gupta \\
  BITS Pilani \\
  \texttt{mailguptakshitij@gmail.com} 
  }

\begin{document}
\maketitle
\begin{abstract}
We present a novel data set, \textsc{WhoDunIt}, to assess the deductive reasoning capabilities of large language models (LLM) within narrative contexts. 
Constructed from open domain mystery novels and short stories, the dataset challenges LLMs to identify the perpetrator after reading and comprehending the story. 
To evaluate model robustness, we apply a range of character-level name augmentations, including original names, name swaps, and substitutions with well-known real and/or fictional entities from popular discourse.
We further use various prompting styles to investigate the influence of prompting on deductive reasoning accuracy.

We conduct evaluation study with state-of-the-art models, specifically \textit{GPT-4o, GPT-4-turbo,} and \textit{GPT-4o-mini}, evaluated through multiple trials with majority response selection to ensure reliability. 
The results demonstrate that while LLMs perform reliably on unaltered texts, accuracy diminishes with certain name substitutions, particularly those with wide recognition.
This dataset is publicly available \href{https://huggingface.co/datasets/kjgpta/WhoDunIt}{here}.
\end{abstract}

\section{Introduction}

Large Language Models (LLMs) have demonstrated exceptional capabilities in a wide array of natural language tasks, from text generation and summarization to complex reasoning and inference \citep{brown2020language}. 
The release of the transformer architecture by \citet{vaswani2017attention} marked a pivotal advancement in the field, enabling models to handle long-range dependencies in text more effectively through self-attention mechanisms. 
This breakthrough not only enhanced model scalability but also laid the foundation for the development of increasingly sophisticated LLMs that are now capable of handling nuanced and context-rich tasks. 
With the emergence of models such as BERT\citep{kenton2019bert}, GPT-2\citep{radford2019language}, and later ChatGPT\citep{openai_chatgpt}, the field of natural language processing has seen rapid innovation, driving significant improvements in model performance and expanding potential applications.

ChatGPT demonstrated that LLMs could deliver highly interactive, contextually relevant responses in real-time, broadening their accessibility to non-technical users and sparking widespread integration in industries. 
This release emphasized the need for systematic evaluation frameworks to understand the capabilities, limitations, and potential biases of these models as they are adopted in real-world applications.

Over recent years, several significant benchmarks have been introduced, such as MMLU\citep{hendrycks2020measuring}, HELM\citep{liang2022holistic}, 
Open LLM Leaderboard\footnote{\url{https://huggingface.co/spaces/open-llm-leaderboard/open_llm_leaderboard}}, and AlpacaEval\footnote{\url{https://github.com/tatsu-lab/alpaca_eval}}. 
These benchmarks have been critical in capturing LLM reasoning capabilities and enabling comparisons among state-of-the-art models.

This paper contributes to these efforts by introducing a novel dataset specifically designed to assess deductive reasoning within narrative contexts. 
To build this dataset we take inspiration from a recent interview\cite{huang_sutskever_interview_2023} between Ilya Sutskever and Jensen Huang about "next word prediction" being sufficient for understanding.
Our benchmark aims to provide deeper insights into the adaptability and inference capabilities of leading models, including \textit{GPT-4o, GPT-4-turbo,} and \textit{GPT-4o-mini} \citep{gpt4}, especially in tasks involving complex narrative comprehension.
We believe that such a benchmark will help future model iteration on LLMs deductive reasoning capabilities as well as complex long-form narrative comprehension. 


This paper is organized as follows: Section 2 reviews relevant prior research, while Section 3 details the dataset preparation process. 
In Section 4, we describe the experimental setup used for evaluation. 
Section 5 presents our findings and analyzes them in terms of LLM capabilities. 
Finally, Section 6 offers conclusions and outlines directions for future work.

\section{Related Works}

Foundational LLMs, such as GPT-2 and GPT-3, demonstrated strong performance across various text-based tasks, though they initially struggled with complex, multi-step reasoning \citep{radford2019language, brown2020language}.


CoT prompting, which encourages models to break down problems into logical steps, has been shown to enhance accuracy and coherence in deductive tasks \citep{wei2022chain}. 
Additional methods, like Self-Reflection prompting, further improve reliability by having models verify and refine their responses, leading to more thoughtful answers \citep{shinn2024reflexion, madaan2024self}.


LLMs' abilities to handle narrative reasoning—tracking characters, plot progression, and thematic elements—have also been a focal area of AI research. 
Studies have shown that while models can generate coherent stories, they often struggle with consistency over long narratives \citep{ammanabrolu2021automated, rashkin2020plotmachines}. 
Enhanced approaches have aimed to improve narrative coherence, though challenges remain, particularly in maintaining character roles and logical plot flow.


Several benchmarks assess LLMs’ reasoning and comprehension, including MMLU, HELM, and Big-Bench (BBH), which evaluate performance across diverse tasks \citep{hendrycks2020measuring, liang2022holistic, srivastava2022beyond}. 
These benchmarks incorporate tasks requiring reasoning and narrative comprehension, though few focus specifically on deductive reasoning within mystery narratives.

\section{Dataset Preparation}

\begin{figure}[h]
    \flushright 
    \includegraphics[width=0.5\textwidth]{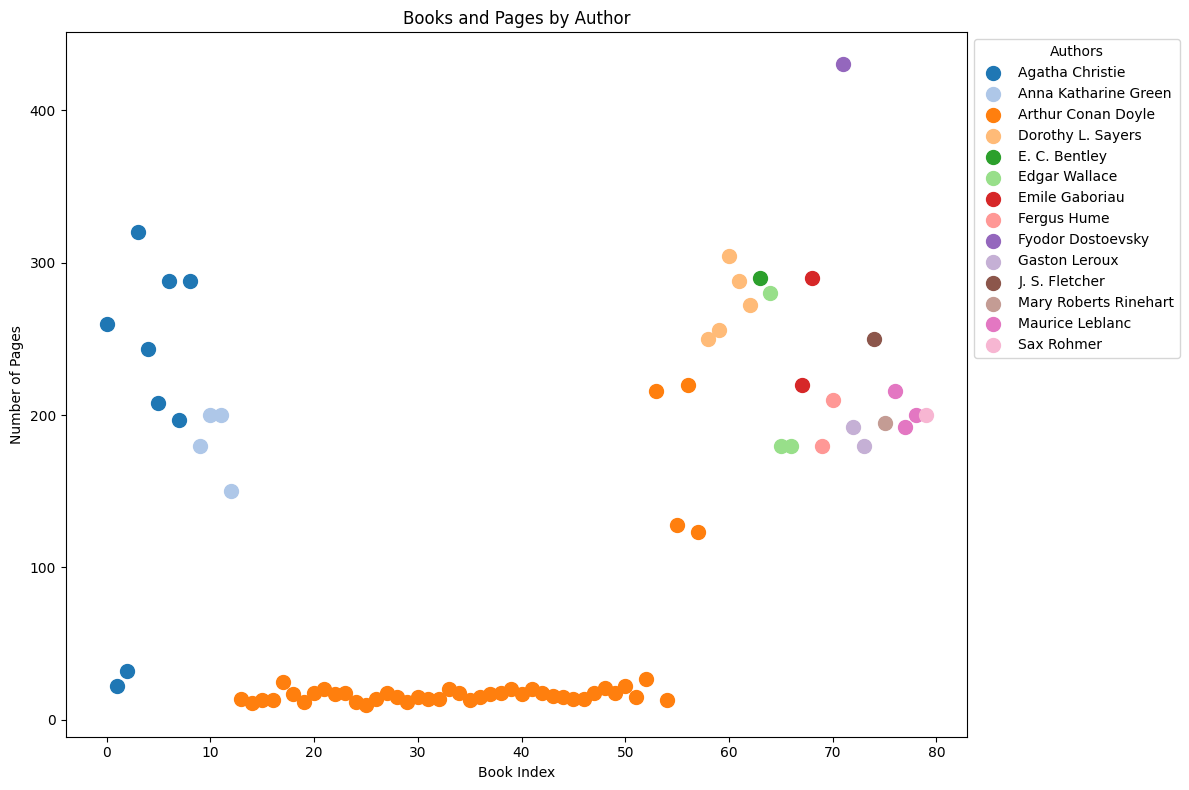} 
    \caption{Distribution by Length}
    \label{fig:author_pages}
\end{figure}

In this section, we outline our dataset preparation, validation process. 
To release this dataset for open source use, we focus on books that have entered the public domain, so we use \textit{Project Gutenberg}\footnote{\url{https://www.gutenberg.org/}} as our primary story source. 
We then obtained the list of 500+ Mystery and Detective story titles, that are of interest to us. 
Additionally, to maintain sufficient variability and diversity in the dataset, we ensured that we represent all the broad characteristics of the stories.
Each selected novel features an identifiable culprit, ensuring that the task involves pinpointing to perpetrator.
The novels span a diverse range of authors and storytelling styles, encompassing classic \textit{WhoDunIt} detective novels by authors such as Agatha Christie. 
As shown in Figure~\ref{fig:author_pages}, the stories vary in length, covering short, medium and full narratives, providing a broad spectrum of text.
By including works from different writers and narrative traditions, we ensure that the models encounter a variety of narrative structures, reasoning styles, and linguistic expressions used to describe mystery and crime.

Since these stories are very popular and have been in the public discourse for a long time, for most of the stories, we find the identity of the culprit from services like \textit{Cliffnotes}\footnote{\url{https://www.cliffsnotes.com/}}. 
This provides us confidence about the identity of the culprit of the story, and hence the accuracy of our dataset. 
Secondly, for others we read them ourselves to figure out the culprit of the story. 





Since these stories are in public domain, any model has most likely already been trained on them. 
Additionally, model would also have trained on any notes/blog posts about these stories. 
Thus the identity of culprit is probably already in model's memory. 
To further investigate whether the model depends on memorized data from pre-training or can genuinely engage in contextual reasoning, we applied a series of character-name substitutions. 
Each augmentation is intended to disrupt potential memorized associations with names, forcing the model to rely on contextual cues and relationships between characters, rather than merely recognizing famous names. 

Here are the specific augmentations and the rationale behind each:

\begin{itemize}
    \item \textbf{Original Character Names:} This serves as a control, where no modifications are made to the text, providing a baseline for the model’s deduction capabilities with familiar, unchanged names.
    \item \textbf{Full Character Name Swap:} Here, we swap the names of all characters in the story. 
This approach is intended to test the model’s capacity to follow complex character interactions and relationships without relying on the original names. 
This alteration simulates a scenario where familiar identifiers are altered, requiring the model to deduce based on narrative function rather than name recognition.
    \item \textbf{Replacement with Harry Potter Character Names:} In this augmentation, we replace all character names with those of well-known characters from the Harry Potter series. 
This tactic tests the model’s ability to ignore pre-trained associations tied to widely recognized fictional characters, focusing instead on the plot’s internal logic and character roles within the story.
    \item \textbf{Hollywood Celebrity Names:} Replacing names with those of famous Hollywood celebrities introduces a real-world layer of familiarity, which can potentially interfere with the model’s reasoning if it relies on pre-trained biases. 
This approach assesses the model’s ability to disregard prominent, real-world associations and concentrate solely on the characters’ roles within the narrative structure.
    \item \textbf{Bollywood Celebrity Names:} Similarly, substituting names with Bollywood celebrities introduces an additional layer of cultural recognition. 
This augmentation not only adds diversity to the test but also evaluates whether the model can apply the same deductive process across different cultural references, further examining its adaptability and robustness under diverse, globally recognizable identities.
\end{itemize}

By applying these augmentation techniques, we systematically modify the dataset to create various degrees of reasoning difficulty, thus challenging the LLM’s deductive capabilities in unique ways. 
Each augmentation serves to disrupt familiar name associations, encouraging the model to prioritize contextual understanding and narrative roles over memorized patterns or recognizable identities.



The list of novels used can be found in the Appendix \ref{appendix:table:novels} and few examples of the point of reveal in stories of our dataset\footnote{\url{https://huggingface.co/datasets/kjgpta/WhoDunIt}} can be found in the Appendix \ref{appendix:dataset_story1}, \ref{appendix:dataset_story2}.

\section{Experimental Setup}

We conducted our experiments on three OpenAI models: \textit{GPT-4o, GPT-4-turbo,} and \textit{GPT-4o-mini} \citep{gpt4}, using OpenAI’s Batch API\footnote{\url{https://platform.openai.com/docs/guides/batch/overview}} via the chat-completions endpoint. 
These models represent a spectrum of capabilities within the GPT-4 family, allowing us to examine how model size and design impact performance in narrative deduction tasks.



    


\subsection{Prompting Techniques}
To assess the models’ reasoning abilities, we applied four prompting styles:

\begin{enumerate}
    \item \textbf{Basic Prompting}: Basic prompting without additional guidance, providing a baseline for model performance \citep{brown2020language}.
    
    \item \textbf{Self-Reflection Prompting}: The model is encouraged to review its response for accuracy, simulating a reflective process that can improve answer quality \citep{shinn2024reflexion}.
    
    \item \textbf{Chain-of-Thought(CoT) Prompting}: Instructs the model to reason through tasks step-by-step, enhancing clarity and accuracy in complex problem-solving \citep{wei2022chain}.
    
    \item \textbf{CoT + Self-Reflection}: Combines step-by-step reasoning with self-reflection, prompting the model to refine its answer after an initial response for improved reliability \citep{madaan2024self}.
\end{enumerate}

To reduce the variability of responses, and ensure we capture the maximum level of LLM reasoning, we consider a 10-shot prompting for each prompt variety and use the most frequent response as the answer\citep{wang2022self}.

With basic prompt as baseline, the self-reflexion is better than that signifying that reflective check fairly improves the performance and adding COT to both of these add fairly to the accuracy of the system.

\section{Results and Analysis}

To ensure robust and reliable results, we evaluated each model’s performance by conducting 10 independent calls for each configuration\citep{wang2022self}. 
In each trial, we maintained consistent input conditions—specifically, the same story, augmentation technique, and prompting style. 
This multi-call approach enabled us to assess the stability and accuracy of each model’s outputs under identical conditions, providing a solid basis for comparative analysis across different model setups.

\subsection{Model Comparison}

\begin{figure}[h]
    \includegraphics[width=0.5\textwidth]{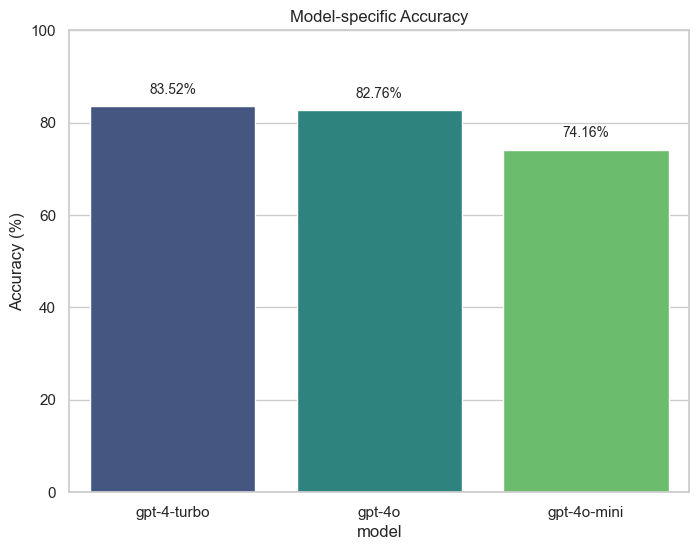}
    \caption{Accuracy comparison across models}
    \label{fig:model_accuracy}
\end{figure}

The \textit{GPT-4-turbo} and \textit{GPT-4o} model demonstrated similar high accuracies of 83.5\% and 82.7\%, respectively, showcasing their robust capabilities in handling reasoning tasks. The \textit{GPT-4o-mini}, while smaller, achieved an accuracy of 74.1\%, indicating its proficiency despite having fewer parameters. Figure~\ref{fig:model_accuracy} summarizes the accuracy of each model across different configurations, highlighting the comparable performance of \textit{GPT-4-turbo} and \textit{GPT-4o} due to their advanced reasoning and inference abilities.

\subsection{Impact of Document Length on Model Accuracy}
\begin{figure}[h]
\centering
\includegraphics[width=0.5\textwidth]{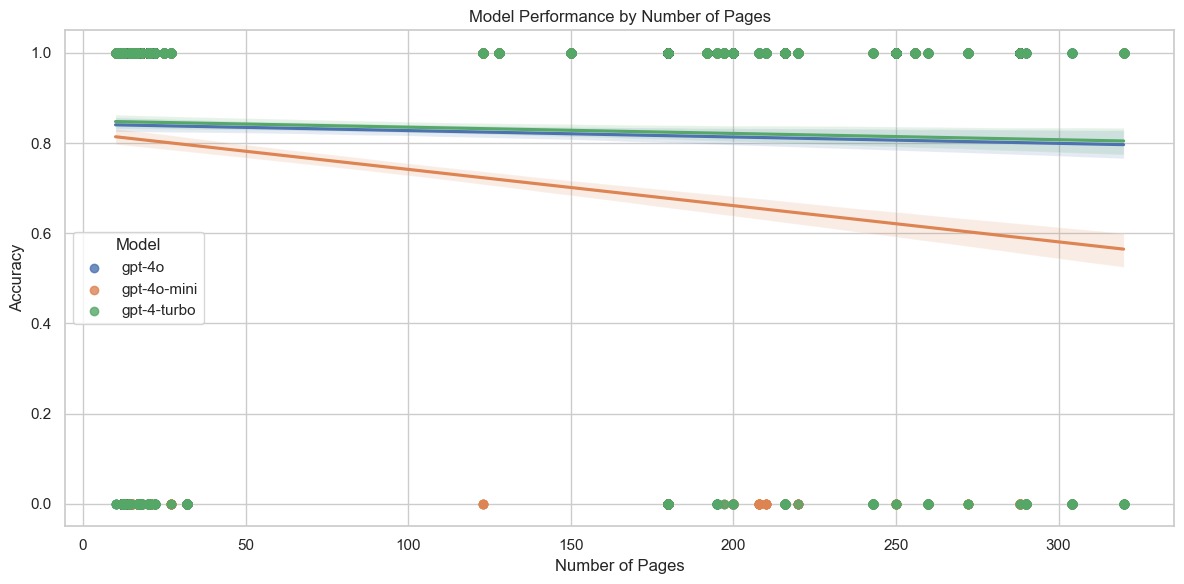}
\caption{Accuracy distribution across the number of pages for different models.}
\label{fig:pages}
\end{figure}

Figure~\ref{fig:pages} demonstrates how model accuracy is influenced by the number of pages in a document. The results indicate that \textit{gpt-4o} and \textit{gpt-4-turbo} exhibit strong resilience to increasing document lengths, maintaining consistent accuracy with only a minor decline as the number of pages grows. This suggests that these models are better equipped to handle long-context scenarios without significant performance degradation.

On the other hand, \textit{gpt-4o-mini} shows a pronounced decline in accuracy as the number of pages increases. This steep drop-off highlights its limitations in processing and retaining information in longer documents. The disparity between \textit{gpt-4o-mini} and the other models becomes more evident as the document length increases.


\subsection{Data Augmentation Analysis}

The models achieved similar highest accuracy on the original text. However, when all character names were swapped, there was a noticeable drop in accuracy, suggesting that extensive alterations to familiar name patterns hinder the model's understanding of the narrative.

\begin{figure}[h]
\centering
\includegraphics[width=0.5\textwidth]{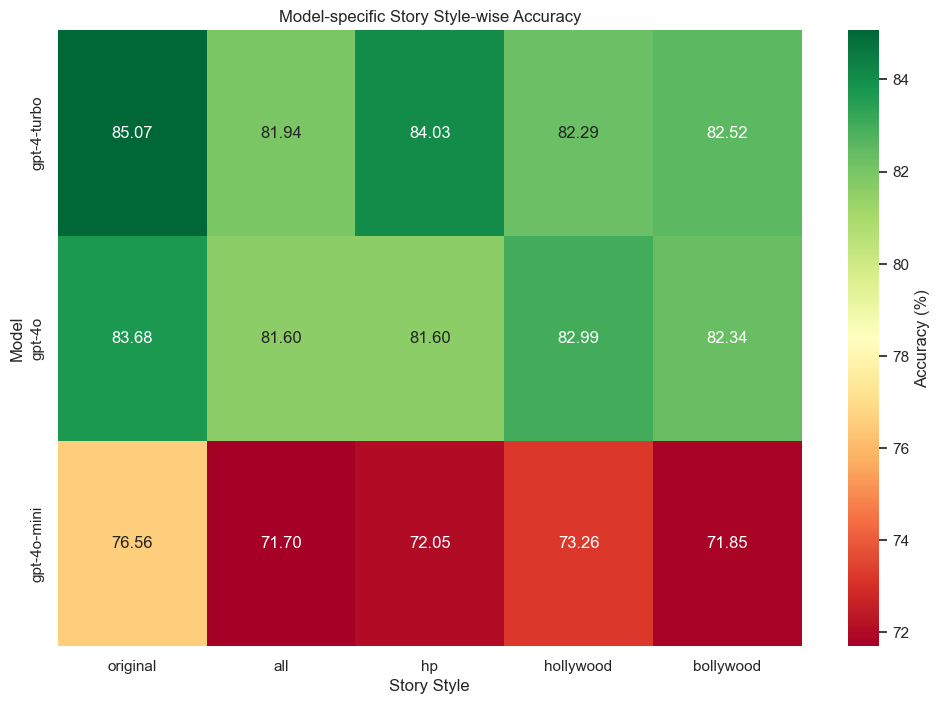}
\caption{Accuracy across different data augmentation techniques.}
\label{fig:augmentation_accuracy}
\end{figure}

Interestingly, the accuracy increased for the \textit{Harry Potter,} \textit{Hollywood,} and \textit{Bollywood} versions of the text, with the model performing similarly across these three cases. This indicates that the model benefits from contexts associated with well-known entities, possibly due to pre-training on a large corpus containing such references. Figure~\ref{fig:augmentation_accuracy} summarizes the accuracy of each text variation, highlighting how character name familiarity and context influence model performance.



The table below specifies the meaning of different augmentation styles used in the analysis:

\begin{table}[h!]
\centering
\resizebox{0.5\textwidth}{!}{%
\begin{tabular}{|m{3cm}|m{8cm}|}
\hline
\textbf{Story Style} & \textbf{Description} \\ \hline\hline
\textbf{original}             & Original text without any alterations. \\ \hline
\textbf{all}                  & All character names in the story swapped. \\ \hline
\textbf{hp}                   & Story with Harry Potter theme augmentation. \\ \hline
\textbf{hollywood}            & Story augmented with a Hollywood theme. \\ \hline
\textbf{bollywood}            & Story augmented with a Bollywood theme. \\ \hline
\end{tabular}%
}
\caption{Descriptions of different augmentation styles.}
\label{tab:story_styles}
\end{table}

\subsection{Prompting Technique Analysis}

Prompting techniques had a notable impact on the model’s ability to deduce the culprit’s identity, with each method contributing differently to accuracy.

\begin{itemize}
    \item \textbf{Normal Prompting}: As a baseline, normal prompting resulted in a relatively lower precision, as the model produced direct responses without deeper reasoning \citep{brown2020language}.

    \item \textbf{Self-Reflection Prompting}: Accuracy improved with Self-Reflection prompting, where the model refined responses through internal checks, leading to greater consistency in deductions \citep{shinn2024reflexion}.

    \item \textbf{Chain-of-Thought (CoT) Prompting}: CoT prompting further increased accuracy by guiding the model through a structured reasoning process, allowing it to systematically address key narrative elements \citep{wei2022chain}.

    \item \textbf{Chain-of-Thought + Self-Reflection (CoT + Self-Reflection)}: The combination of CoT and Self-Reflection yielded similar results as CoT, as the model generated logical step-by-step responses and then refined them, demonstrating the enhanced performance in narrative deduction \citep{madaan2024self}.
\end{itemize}

\begin{figure}[h]
\centering
\includegraphics[width=0.5\textwidth]{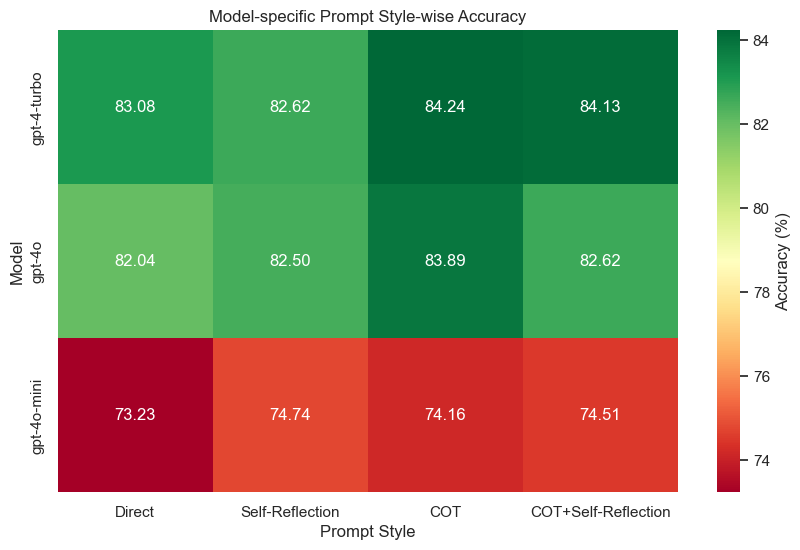}
\caption{Accuracy across different prompting techniques.}
\label{fig:prompting_accuracy}
\end{figure}

Figure \ref{fig:prompting_accuracy} presents the accuracy achieved by each prompting technique, with substantial gains observed by adding CoT and Self-Reflexion, underscoring the effectiveness of combining structured reasoning and reflective validation.

Our results \footnote{Code and results are available on \href{https://github.com/kjgpta/WhoDunIt-Evaluation_benchmark_for_culprit_detection_in_mystery_stories}{GitHub}} reveal that model architecture, data augmentation, and prompting techniques all play a significant role in shaping deductive performance. 
The findings highlight the crucial impact of structured prompting  on enhancing model accuracy, particularly in complex narrative deduction tasks. 
These insights underscore the need for refined prompting strategies and comprehensive data preparation to optimize LLMs capabilities in inference-driven applications.

\section{Conclusion and Future Work}

We conclude by releasing our deductive reasoning capability benchmark, called \textsc{WhoDunIt}.
We use this dataset to examine the deductive reasoning capabilities of large language models (LLMs) in complex narrative contexts, specifically focusing on mystery narratives that require nuanced inference and multi-step reasoning. 
Using a structured evaluation framework, we assessed the effects of model architecture, data augmentation, and various prompting techniques on the deductive accuracy of these LLM configurations — \textit{GPT-4o, GPT-4-turbo,} and \textit{GPT-4o-mini}. 
Our findings indicate that a combination of structured reasoning and reflective validation techniques, namely Chain-of-Thought and Self-Reflection prompting, significantly enhances model performance. 
Our results indicate that before a detective level reasonable understanding the models still have some progress to go in long-form narrative comprehension, and have to build robustness to changes in character names, while keeping the story plot intact. 
A key aspect of future work would be building long form comprehensive puzzle dataset, that would be able to test the limits of the LLM reasoning capabilities, and to reduce the impact of bias inducted during pre-training. 

\section*{Limitations}
This study is limited to short and medium-length stories due to the model's context length constraints, which restrict the analysis of longer narratives.

\bibliography{custom}
\newpage
\appendix
\section{Appendix} 

\subsection{Extract from \textit{A Case of Identity} by Arthur Conan Doyle}\label{appendix:dataset_story1}

\textbf{\textit{Culprit:} James Windibank}

\textbf{\textit{Point of Reveal:}}
\begin{quote}
“My dear fellow,” said Sherlock Holmes as we sat on either side of the fire in his lodgings at Baker Street, “life is infinitely stranger than anything which the mind of man could invent.
We would not dare to conceive the things which are really mere commonplaces of existence.

...

"Certainly," said Holmes, stepping over and turning the key in the door. "I let you know, then, that I have caught him!"

"What! where?" shouted Mr. Windibank, turning white to his lips and glancing about him like a rat in a trap.

\textcolor{red}{\textbf{"Oh, it won’t do—really it won’t," said Holmes suavely. "There is no possible getting out of it, Mr. Windibank. It is quite too transparent, and it was a very bad compliment when you said that it was impossible for me to solve so simple a question. That’s right! Sit down and let us talk it over."}}

Our visitor collapsed into a chair, with a ghastly face and a glitter of moisture on his brow. "It—it’s not actionable," he stammered.

...

As I expected, his reply was typewritten and revealed the same trivial but characteristic defects. 
The same post brought me a letter from Westhouse \& Marbank, of Fenchurch Street, to say that the description tallied in every respect with that of their employe, James Windibank. 
Voila tout” “And Miss Sutherland?” “If I tell her she will not believe me. You may remember the old Persian saying, ‘There is danger for him who taketh the tiger cub, and danger also for whoso snatches a delusion from a woman.’ There is as much sense in Hafiz as in Horace, and as much knowledge of the world.”

\end{quote}

\subsection{Extract from \textit{Silver Blaze} by Arthur Conan Doyle}\label{appendix:dataset_story2}

\textbf{\textit{Culprit:} John Straker}

\textbf{\textit{Point of Reveal:}}
\begin{quote}
I am afraid, Watson, that I shall have to go,” said Holmes, as we sat down together to our breakfast one morning. “Go! Where to?” “To Dartmoor; to King’s Pyland.”

...

“The real murderer is standing immediately behind you.” He stepped past and laid his hand upon the glossy neck of the thoroughbred.

“The horse!” cried both the Colonel and myself.

\textcolor{red}{\textbf{“Yes, the horse. And it may lessen his guilt if I say that it was done in self-defence, and that John Straker was a man who was entirely unworthy of your confidence. But there goes the bell, and as I stand to win a little on this next race, I shall defer a lengthy explanation until a more fitting time.”}}

...

 My eyes fell upon the sheep, and I asked a question which, rather to my surprise, showed that my surmise was correct. 
“When I returned to London I called upon the milliner, who had recognised Straker as an excellent customer of the name of Derbyshire, who had a very dashing wife, with a strong partiality for expensive dresses. 
I have no doubt that this woman had plunged him over head and ears in debt, and so led him into this miserable plot.” “You have explained all but one thing,” cried the Colonel
“Where was the horse?” “Ah, it bolted, and was cared for by one of your neighbours. 
We must have an amnesty in that direction, I think. This is Clapham Junction, if I am not mistaken, and we shall be in Victoria in less than ten minutes. 
If you care to smoke a cigar in our rooms, Colonel, I shall be happy to give you any other details which might interest you.
\end{quote}
\subsection{List of Stories and Authors}\label{appendix:table:novels}
\onecolumn
\begin{longtable}[H]{|p{2cm}|p{7cm}|p{4cm}|}
\hline
\textbf{Type} & \textbf{Title} & \textbf{Author Name} \\ \hline
Novel & A Study in Scarlet & Arthur Conan Doyle \\ \hline
Novel & Crime and Punishment & Fyodor Dostoevsky \\ \hline
Novel & Clouds of Witness & Dorothy L. Sayers \\ \hline
Novel & File No. 113 & Emile Gaboriau \\ \hline
Novel & Find the Woman & G. K. Chesterton \\ \hline
Novel & Silver Blaze & Arthur Conan Doyle \\ \hline
Novel & That Affair Next Door & Anna Katherine Green \\ \hline
Novel & The Borough Treasurer & J. S. Fletcher \\ \hline
Novel & The Clue of the Twisted Candle & Edgar Wallace \\ \hline
Novel & The Crooked Man & Arthur Conan Doyle \\ \hline
Novel & The Crystal Stopper & Maurice Leblanc \\ \hline
Novel & The Curved Blades & Carolyn Wells \\ \hline
Novel & The D'Arblay Mystery & R. Austin Freeman \\ \hline
Novel & The Fellowship of the Frog & Edgar Wallace \\ \hline
Novel & The Hound of the Baskervilles & Arthur Conan Doyle \\ \hline
Novel & The Insidious Dr. Fu Manchu & Sax Rohmer \\ \hline
Novel & The Leavenworth Case & Anna Katherine Green \\ \hline
Novel & The Lerouge Case & Emile Gaboriau \\ \hline
Novel & The Man in Lower Ten & Mary Roberts Rinehart \\ \hline
Novel & The Man in the Brown Suit & Agatha Christie \\ \hline
Novel & The Murder of Roger Ackroyd & Agatha Christie \\ \hline
Novel & The Murder on the Links & Agatha Christie \\ \hline
Novel & The Mysterious Affair at Styles & Agatha Christie \\ \hline
Novel & The Mystery of the Blue Train & Agatha Christie \\ \hline
Novel & The Mystery of the Yellow Room & Gaston Leroux \\ \hline
Novel & The Opal Serpent & Fergus Hume \\ \hline
Novel & The Problem of Thor Bridge & Arthur Conan Doyle \\ \hline
Novel & The Secret Adversary & Agatha Christie \\ \hline
Novel & The Sign of the Four & Arthur Conan Doyle \\ \hline
Novel & The Teeth of the Tiger & Maurice Leblanc \\ \hline
Novel & The Unpleasantness at the Bellona Club & Dorothy L. Sayers \\ \hline
Novel & The Valley of Fear & Arthur Conan Doyle \\ \hline
Novel & Trent's Last Case & E. C. Bentley \\ \hline
Novel & Unnatural Death & Dorothy L. Sayers \\ \hline
Novel & Whose Body? A Lord Peter Wimsey Novel & Dorothy L. Sayers \\ \hline
Novel & X Y Z: A Detective Story & Anna Katherine Green \\ \hline
Short Story & A Case of Identity & Arthur Conan Doyle \\ \hline
Short Story & Silver Blaze & Arthur Conan Doyle \\ \hline
Short Story & The Adventure of Black Peter & Arthur Conan Doyle \\ \hline
Short Story & The Adventure of Charles Augustus Milverton & Arthur Conan Doyle \\ \hline
Short Story & The Adventure of Shoscombe Old Place & Arthur Conan Doyle \\ \hline
Short Story & The Adventure of the Abbey Grange & Arthur Conan Doyle \\ \hline
Short Story & The Adventure of the Beryl Coronet & Arthur Conan Doyle \\ \hline
Short Story & The Adventure of the Blue Carbuncle & Arthur Conan Doyle \\ \hline
Short Story & The Adventure of the Bruce-Partington Plans & Arthur Conan Doyle \\ \hline
Short Story & The Adventure of the Cardboard Box & Arthur Conan Doyle \\ \hline
Short Story & The Adventure of the Copper Beeches & Arthur Conan Doyle \\ \hline
Short Story & The Adventure of the Creeping Man & Arthur Conan Doyle \\ \hline
Short Story & The Adventure of the Dancing Men & Arthur Conan Doyle \\ \hline
Short Story & The Adventure of the Devil's Foot & Arthur Conan Doyle \\ \hline
Short Story & The Adventure of the Dying Detective & Arthur Conan Doyle \\ \hline
Short Story & The Adventure of the Egyptian Tomb & Agatha Christie \\ \hline
Short Story & The Adventure of the Engineer’s Thumb & Arthur Conan Doyle \\ \hline
Short Story & The Adventure of the Empty House & Arthur Conan Doyle \\ \hline
Short Story & The Adventure of the Final Problem & Arthur Conan Doyle \\ \hline
Short Story & The Adventure of the Golden Pince-Nez & Arthur Conan Doyle \\ \hline
Short Story & The Adventure of the Illustrious Client & Arthur Conan Doyle \\ \hline
Short Story & The Adventure of the Mazarin Stone & Arthur Conan Doyle \\ \hline
Short Story & The Adventure of the Norwood Builder & Arthur Conan Doyle \\ \hline
Short Story & The Adventure of the Priory School & Arthur Conan Doyle \\ \hline
Short Story & The Adventure of the Red Circle & Arthur Conan Doyle \\ \hline
Short Story & The Adventure of the Second Stain & Arthur Conan Doyle \\ \hline
Short Story & The Adventure of the Six Napoleons & Arthur Conan Doyle \\ \hline
Short Story & The Adventure of the Solitary Cyclist & Arthur Conan Doyle \\ \hline
Short Story & The Adventure of the Speckled Band & Arthur Conan Doyle \\ \hline
Short Story & The Adventure of the Sussex Vampire & Arthur Conan Doyle \\ \hline
Short Story & The Adventure of the Three Gables & Arthur Conan Doyle \\ \hline
Short Story & The Adventure of the Three Garridebs & Arthur Conan Doyle \\ \hline
Short Story & The Adventure of Wisteria Lodge & Arthur Conan Doyle \\ \hline
Short Story & The Boscombe Valley Mystery & Arthur Conan Doyle \\ \hline
Short Story & The Disappearance of Lady Frances Carfax & Arthur Conan Doyle \\ \hline
Short Story & The Five Orange Pips & Arthur Conan Doyle \\ \hline
Short Story & The Hunter's Lodge Case & Agatha Christie \\ \hline
Short Story & The Musgrave Ritual & Arthur Conan Doyle \\ \hline
Short Story & The Naval Treaty & Arthur Conan Doyle \\ \hline
Short Story & The Red-Headed League & Arthur Conan Doyle \\ \hline
Short Story & The Riddle of the Purple Emperor & Fergus Hume \\ \hline
Short Story & The Sturgis Wager: A Detective Story & Anna Katherine Green \\ \hline
\end{longtable}
\twocolumn

\end{document}